\documentclass{article}

\usepackage{times}
\usepackage[dvipdfmx]{graphicx} 
\usepackage{subfigure} 

\usepackage{natbib}

\usepackage{algorithm}
\usepackage{algorithmic}

\usepackage{hyperref}

\usepackage[accepted]{whi2016}

\usepackage[cmex10]{amsmath}
\usepackage{amssymb}
\usepackage{amsfonts}
\usepackage{url}
\usepackage{bm}
\usepackage{multirow}
\usepackage{array}

\newtheorem{problem}{Problem}

\newcommand{\R}{\ensuremath{\mathbb{R}}}

\newcommand{\vi}{{\vert\kern-0.25ex\vert\kern-0.25ex\vert}}

\icmltitlerunning{Making Tree Ensembles Interpretable}

\begin{document} 

\twocolumn[
\icmltitle{Making Tree Ensembles Interpretable}

\icmlauthor{Satoshi Hara}{satohara@nii.ac.jp}
\icmladdress{National Institute of Informatics, Japan\\
JST, ERATO, Kawarabayashi Large Graph Project}
\icmlauthor{Kohei Hayashi}{hayashi.kohei@gmail.com}
\icmladdress{National Institute of Advanced Industrial Science and Technology, Japan}

\icmlkeywords{boring formatting information, machine learning, ICML}

\vskip 0.3in
]

\begin{abstract} Tree ensembles, such as random forest and boosted trees, are renowned for their high prediction performance, whereas their interpretability is critically limited.
In this paper, we propose a post processing method that improves the model interpretability of tree ensembles.
After learning a complex tree ensembles in a standard way, we approximate it by a simpler model that is interpretable for human.
To obtain the simpler model, we derive the EM algorithm minimizing the KL divergence from the complex ensemble. 
A synthetic experiment showed that a complicated tree ensemble was approximated reasonably as interpretable.
\end{abstract} 

\section{Introduction}
\label{sec:intro}

Ensemble models of decision trees such as random forests~\cite{breiman2001random} and boosted trees~\cite{friedman2001greedy} are popular machine learning models, especially for prediction tasks. Because of their high prediction performance, they are one of the must-try methods when dealing with real problems.
Indeed, they have attained high scores in many data mining competitions such as web ranking~\cite{mohan2011web}.
These tree ensembles are collectively referred to as \textit{Additive Tree Models} (ATMs)~\cite{cui2015optimal}.

A main drawback of ATMs is in interpretability. They divide an input space by a number of small regions and make prediction depending on a region. Usually, the number of the regions they generate is over thousand, which roughly means that there are thousands of different rules for prediction. Non-expert people cannot understand such tremendous number of rules. A decision tree, on the other hand, is well known as one of the most interpretable models. Despite weak prediction ability, the number of regions generated by a single tree is drastically small, which makes the model transparent and understandable.

Obviously, there is a tradeoff between prediction performance and interpretability.  \citet{eto2014fully} proposed a simplification method of a tree model that prunes redundant branches by approximated Bayesian inference. \citet{wang2015trading} studied a similar approach with a richly-structured tree model. Although these approaches certainly improve interpretability, prediction performance is inevitably degenerated, especially when a drastic simplification is needed.

\begin{figure*}[t]
	\centering
	\subfigure[Original Data]{
		\includegraphics[width=0.32\textwidth]{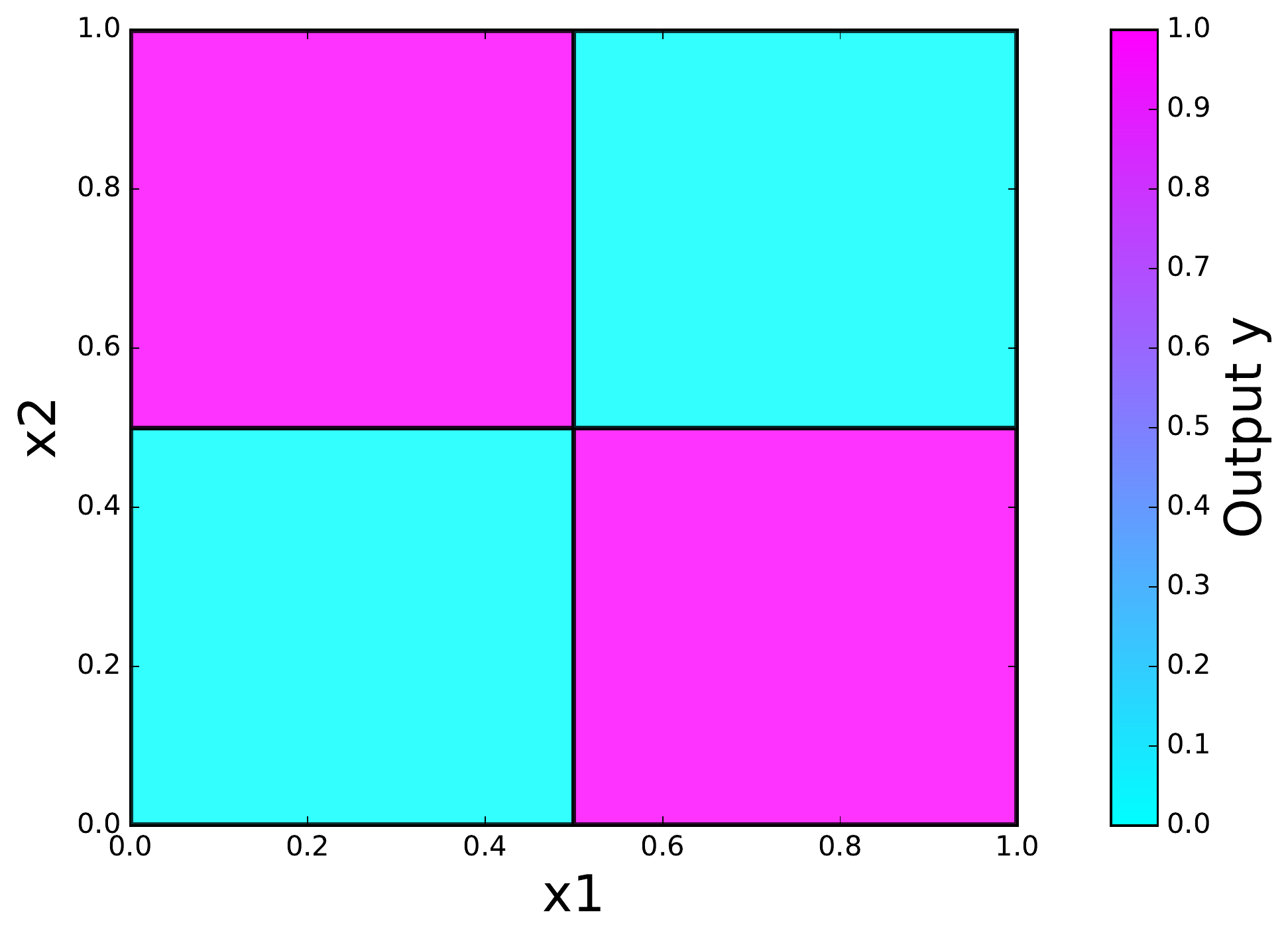}
		\label{fig:true}}
	\hspace{-4pt}
	\subfigure[Learned Tree Ensemble]{
		\includegraphics[width=0.32\textwidth]{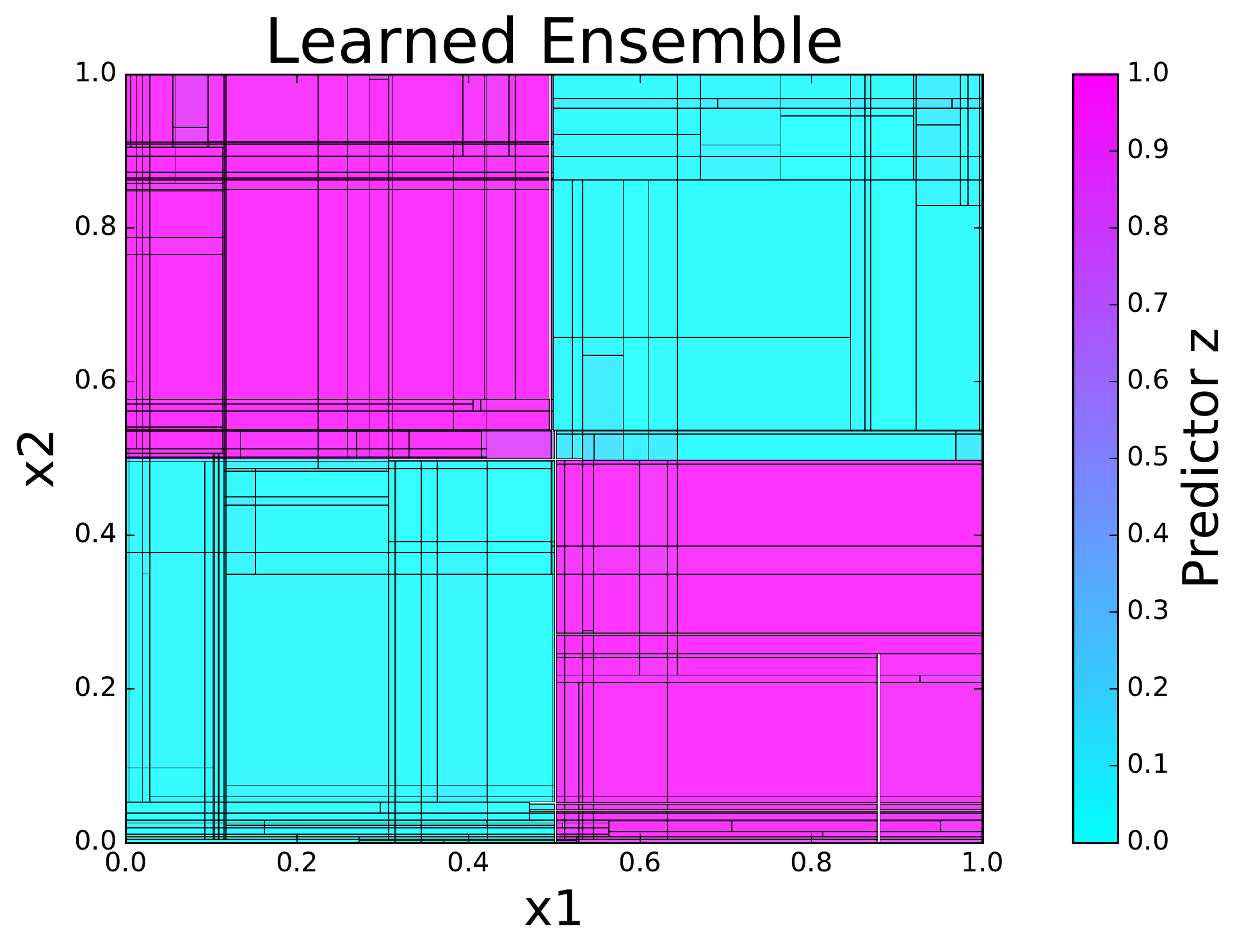}
		\label{fig:before}}
	\hspace{-4pt}
	\subfigure[Simplified Model]{
		\includegraphics[width=0.31\textwidth]{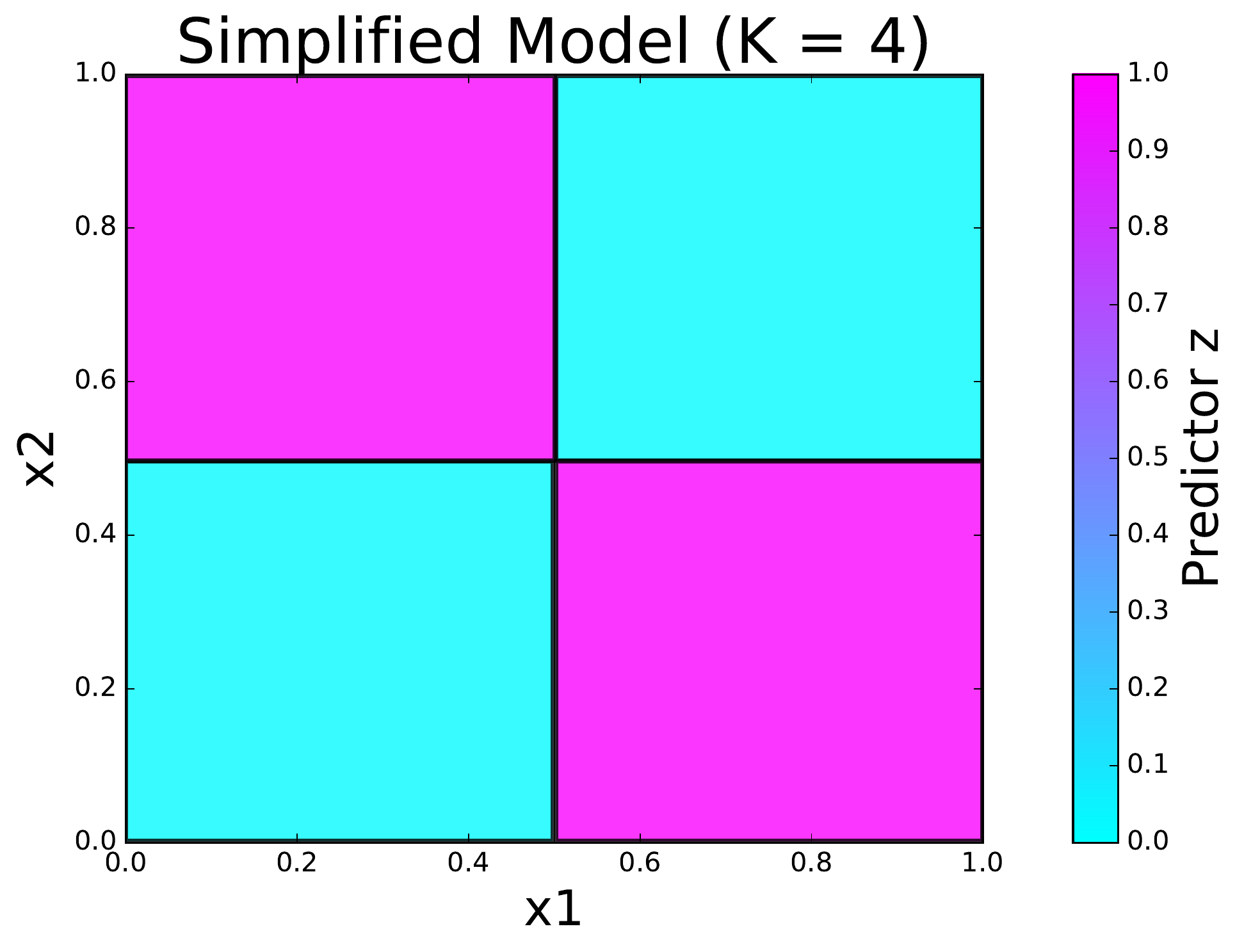} 
		\label{fig:after}}
	\caption{The original data (a) is learned by ATM with 744 regions (b). The complicated ensemble (b) is approximated by four regions using the proposed method (c). Each rectangle shows each input region specified by the model.}
	\label{fig:example}
\end{figure*}

Motivated by the above observation, we study how to improve the interpretability of ATMs. We say an ATM is interpretable if the number of regions is sufficiently small (say, less than ten). Our goal is then formulated as
\begin{enumerate}
\item reducing the number of regions, while
\item minimizing model error.
\end{enumerate}
To satisfy these contradicting requirements, we propose a post processing method. Our method works as follows. 
First, we learn an ATM in a standard way, which generates a number of regions (\figurename~\ref{fig:before}).
Then, we mimic this by a simple model where the number of regions is fixed as small  (\figurename~\ref{fig:after}). 
We refer to the former as the prediction model or \emph{model P}, and the latter as interpretation model or \emph{model I}. 

Our contributions are summarized as follows.

\textbf{Separation of prediction and interpretation.}
We prepare two different models: \emph{model P} for prediction and \emph{model I} for interpretation. This idea balances requirements 1 and 2. 

\textbf{Reformulation of ATMs.}
We reinterpret an ATM as a probabilistic generative model.
With this change of perspective on ATMs, we can model an ATM as a mixture-of-experts~\cite{jordan1994hierarchical}.
In addition, this formulation induces the following optimization algorithm.

\textbf{Optimization algorithm.}
To obtain \emph{model I} that is close to \emph{model P}, we consider the KL divergence between \emph{models I} and \emph{model P}.
To minimize this, we derive the EM algorithm.

\section{Related Work}
\label{sec:rel}

For single tree models, many studies regarding interpretability have been conducted.
One of the most widely used methods would be a decision tree such as CART~\cite{breiman1984classification}.
Oblique decision trees~\cite{murthy1994system} and Bayesian treed linear models~\cite{bernardo2003bayesian} extended the decision tree by replacing single-feature partitioning with linear hyperplane partitioning and the region-wise constant predictor with the linear model.
While using hyperplanes and linear models improve the prediction accuracy, they tend to degrade the interpretability of the models owing to their complex structure.
\citet{eto2014fully} and \citet{wang2015trading} proposed tree-structured mixture-of-experts models.
These models aim to derive interpretable models while maintaining prediction accuracies as much as possible.
We note that all these researches aimed to improve the accuracy--interpretability tradeoff by building a single tree.
In this sense, although they treated the tree models, they are different from our study in the sense that we try to interpret ATMs.

The most relevant study compared with ours would be \emph{inTrees} (interpretable trees) proposed by \citet{deng2014interpreting}.
Their study tackles the same problem as ours, the post-hoc interpretation of ATMs.
The inTrees framework extracts rules from ATMs by treating tradeoffs among the frequency of the rules appearing in the trees, the errors made by the predictions, and the length of the rules.
The fundamental difficulty on inTrees is that its target is limited to the classification ATMs.
Regression ATMs are first transformed into the classification one by discretizing the output, and then inTrees is applied to extract the rules.
The number of discretization level remains as a user tuning parameter, which severely affects the resulting rules.
In contrast, our proposed method can handle both classification and regression ATMs.

\section{ATMs: Additive Tree Models}
\label{sec:atm}

\paragraph{Notation:}
For $N \in \mathbb{N}$, $[N]$ denotes the set of integers $[N] = \{1, 2, \ldots, N\}$.
For a statement $a$, $\mathbb{I}(a)$ denotes the indicator of $a$, i.e. $\mathbb{I}(a) = 1$ if $a$ is true, and $\mathbb{I}(a) = 0$ if $a$ is false.

An ATM is an ensemble of $T$ decision trees.
For simplicity, we consider a simplified version of ATMs.
Let $\bm{x} = (x_1, x_2, ..., x_D) \in \R^D$ be a $D$-dimensional feature vector.
In the paper, we focus on the regression problem where the target domain $\mathcal{Y} = \R$.
Let $f_t(\bm{x})$ be the output from the $t$-th decision tree for an input $\bm{x}$.
The output $z$ of of the ATM is then defined as a weighted sum of all the tree outputs $z = \sum_{t=1}^T \alpha_t f_t(\bm{x})$ with weights $\{\alpha_t\}_{t=1}^T$.

The function $f_t(\bm{x})$, the output from tree $t$, is represented as $f_t(\bm{x}) = \sum_{i_t = 1}^{I_t} z_{i_t} \mathbb{I}(\bm{x} \in R_{i_t})$.
Here, $I_t$ denotes the number of leaves in the tree $t$, $i_t$ denotes the index of the leaf node in the tree $t$, and $R_{i_t}$ denotes the input region specified by the leaf node $i_t$.
Note that the regions are non-overlapping, i.e.\ $R_{i_t} \cap R_{i'_t} = \emptyset$ for $i_t \neq i'_t$.
Using this notation, ATM is written as:
\begin{align}
	\textstyle{z = \sum_{t=1}^T \alpha_t \sum_{i_t=1}^{I_t} z_{i_t} \mathbb{I}(\bm{x} \in R_{i_t})  = \sum_{g=1}^G z_g \mathbb{I}(\bm{x} \in R_g)} ,
    \label{eq:atm}
\end{align}
where $R_g = \cap_{t=1}^T R_{i_t} \neq \emptyset$ for $(i_1, i_2, \ldots, i_T) \in \prod_{t=1}^T [I_t]$ and $z_g = \sum_{t=1}^T \alpha_t z_{i_t}$.

\section{Post-hoc ATM Interpretation Method}
\label{sec:prop}

Recall that our goal is to make an interpretation model \emph{I} by
\begin{enumerate}
	\item reducing the number of regions, while
	\item minimizing model error.
\end{enumerate}
Based on (\ref{eq:atm}), now requirement 1 corresponds to reducing the number of regions G.
Combining this with requirement 2, the problem we want to solve is defined as follows.
\begin{problem}[ATM Interpretation]
	\label{prob:atm}
	Approximate ATM (model P) (\ref{eq:atm}) by using a simplified model (model I) with only $K \ll G$ regions:
	\begin{align*}
		\textstyle{\sum_{g=1}^G z_g \mathbb{I}(\bm{x} \in R_g) \approx \sum_{k=1}^K z'_k \mathbb{I}(\bm{x} \in R'_k)} .
	\end{align*}
\end{problem}
To solve this approximation problem, we need to optimize both the predictors $z'_k$ and the regions $R'_k$.
The difficulty is that $R'_k$ is a non-numeric parameter and it is therefore hard to optimize.

\subsection{Probabilistic Generative Model Expression}

We resolve the difficulty of handling the region parameter $R'_k$ by interpreting ATM as a probabilistic generative model.
For simplicity, we consider the axis-aligned tree structure.
We then adopt the next two modifications.
These modifications can also be extended to oblique decision trees~\cite{murthy1994system}.

\begin{figure}[t]
	\centering
	\includegraphics[width=0.25\textwidth]{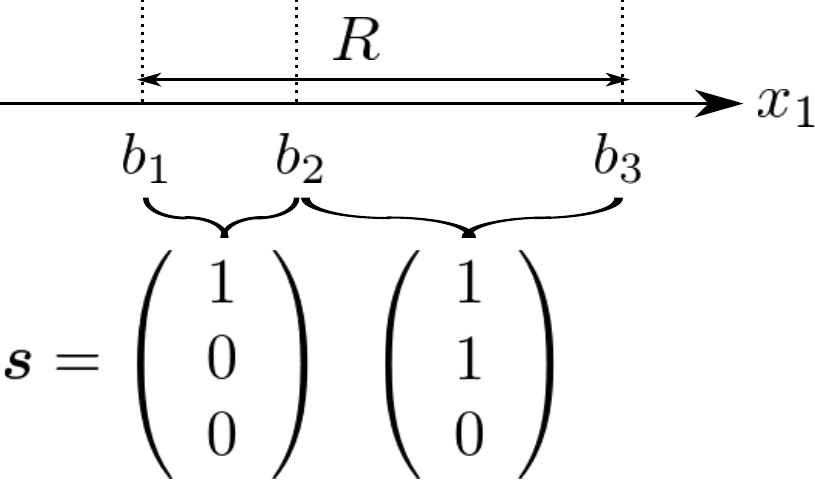}
	\caption{Example of Binary Features. The region $R$ is indicated by the area where $\bm{s}$ matches to a pattern $(1, *, 0)^\top$.}
	\label{fig:region}
\end{figure}

\textbf{Binary Expression of Feature $\bm{x}$:}
Let $\{(d_{j_t}, b_{j_t})\}_{j_t=1}^{J_t}$ be a set of split rules of the tree $t$ where $J_t$ is the number of internal nodes.
At the internal node $j_t$ of the tree $t$, the input is split by the rule $x_{d_{j_t}} < b_{j_t}$ or $x_{d_{j_t}} \geq b_{j_t}$.
Let $\mathcal{L} = \{\{(d_{j_t}, b_{j_t})\}_{j_t=1}^{J_t}\}_{t=1}^T = \{(d_{\ell}, b_{\ell})\}_{\ell=1}^L$ be the set of all the split rules where $L = \sum_{t=1}^T J_t$.
We then define the binary feature $\bm{s} \in \{0, 1\}^L$ by $s_\ell = \mathbb{I}(x_{d_{\ell}} \geq b_{\ell}) \in \{0, 1\}$ for $\ell \in [L]$.
\figurename~\ref{fig:region} shows an example of the binary feature~$\bm{s}$.

\textbf{Generative Model Expression of Regions:}
As shown in \figurename~\ref{fig:region}, one region $R$ can generate multiple binary features.
We can thus interpret the region $R$ as a generative model of a binary feature $\bm{s}$ using a Bernoulli distribution: $p_B(\bm{s} | \bm{\eta}) = \prod_{\ell=1}^L \eta_{\ell}^{s_{\ell}} (1 - \eta_{\ell})^{1 - s_{\ell}}$.
The region $R$ can then be expressed as a pattern of the Bernoulli parameter $\bm{\eta}$:
$\eta_{\ell} = 1$ (or $\eta_{\ell} = 0$) means that the region $R$ satisfies $x_{d_{\ell}} \geq b_{\ell}$ (or $x_{d_{\ell}} < b_{\ell}$), while $0 < \eta_{\ell} < 1$ means that both $x_{d_{\ell}} \geq b_{\ell}$ and $x_{d_{\ell}} < b_{\ell}$ are not relevant to the region $R$.
In the example of \figurename~\ref{fig:region}, the features generated by $R$ are $\bm{s} = (1, 0, 0)^\top$ and $\bm{s} = (1, 1, 0)^\top$, and the Bernoulli parameter is $\bm{\eta} = (1, *, 0)^\top$ where $*$ is in between 0 and 1.

\subsection{Probabilistic Modeling of ATM}

Using the Bernoulli parameter expression of regions, we can express the probability that the predictor $z$ and the binary feature $\bm{s}$ are generated as
\begin{align}
	\textstyle{p(z, \bm{s} | \bm{x}) = \sum_{g=1}^G p(z | g) p(\bm{s} | g) p(g | \bm{x}) } ,
	\label{eq:atmp}
\end{align}
where $p(g | \bm{x}) = \mathbb{I}(\bm{x} \in R_g)$, $p(z | g) = \mathbb{I}(z = z_g)$, and $p(\bm{s} | g) = p_B(\bm{s} | \bm{\eta}_g)$.

To derive \emph{model I}, we approximate the \emph{model P} (\ref{eq:atmp}) using the next mixture-of-experts formulation~\cite{jordan1994hierarchical}:
\begin{align}
	\textstyle{q(z, \bm{s} | \bm{x}) = \sum_{k=1}^K q(z | k) q(\bm{s} | k) q(k | \bm{x})} ,
    \label{eq:q}
\end{align}
where $q(\bm{s} | k) = p_B(\bm{s} | \bm{\eta}_k)$, and
\begin{align*}
	q(k | \bm{x}) &= \frac{\exp (\bm{w}_k^\top \bm{s})}{\sum_{k'=1}^K \exp (\bm{w}_{k'}^\top \bm{s})}, \\
	q(z | k) &= \sqrt{\frac{\lambda_k}{2 \pi}} \exp \left( -\frac{\lambda_k (z - \mu_k)^2}{2} \right) .
\end{align*}
Here, for each $k \in [K]$, $\bm{w}_k \in \R^L$, $\mu_k \in \R$, and $\lambda_k \in \R_+$.
We note that the probabilistic model (\ref{eq:q}) is no longer an ATM but a prediction model with a predictor $\mu_k$ and a set of rules specified by $\bm{\eta}_k$.

\subsection{Parameter Learning with EM Algorithm}

We estimate the parameters of the model (\ref{eq:q}) so that the next KL-divergence to be minimized which solves Problem~\ref{prob:atm}:
\begin{align*}
	\textstyle{\min_{w, \eta, \mu, \lambda} \mathbb{E}_{\hat{p}(\bm{x})}\left[ {\rm KL}[p(z, \bm{s} | \bm{x}) || q(z, \bm{s} | \bm{x})] \right] }, 
\end{align*}
or equivalently, 
\begin{align}
	\textstyle{\max_{w, \eta, \mu, \lambda} \mathbb{E}_{\hat{p}(\bm{x})} \mathbb{E}_{p(z, \bm{s} | \bm{x})} \left[ \log q(z, \bm{s} | \bm{x}) \right] },
	\label{eq:obj}
\end{align}
where $\hat{p}(\bm{x})$ is some input generation distribution,
The finite sample approximation of (\ref{eq:obj}) is
\begin{align}
	\textstyle{ \max_{w, \eta, \mu, \lambda} \sum_{n=1}^N \log q(z^{(n)}, \bm{s}^{(n)} | \bm{x}^{(n)}) }, 
	\label{eq:obj2}
\end{align}
where $\{\bm{x}^{(n)}\}_{n=1}^N \overset{\rm i.i.d.}{\sim} \hat{p}$, and $(z^{(n)}, \bm{s}^{(n)})$ are determined from the input $\bm{x}^{(n)}$.

The optimization problem (\ref{eq:obj2}) is solved by the EM algorithm.
The lower-bound of (\ref{eq:obj2}) is derived as
\begin{align}
	& \textstyle{\sum_{n=1}^N \sum_{k=1}^K \mathbb{E}_{q(u)} [u_k^{(n)}] \left( \log q(z^{(n)} | k) q(\bm{s}^{(n)} | k) q(k | \bm{x}^{(n)})  \right) } \nonumber \\
	& \textstyle{ - \sum_{n=1}^N \sum_{k=1}^K \mathbb{E}_{q(u)} [\log q(u_k^{(n)})] } ,
	\label{eq:lb}
\end{align}
where $u_k^{(n)} \in \{0, 1\}$, $\sum_{k=1}^K u_k^{(n)} = 1$, and $q(u)$ is an arbitrary distribution on $u$~\cite{beal2003variational}.
The EM algorithm is then formulated as alternating maximization with respect to $q$ (E-step) and the parameters (M-step).

\paragraph{[E-Step]}
In E-Step, we fix the values of $w$, $\eta$, $\mu$, and $\lambda$, and we maximize the lower-bound (\ref{eq:lb}) with respect to the distribution $q(u)$, which yields
\begin{align*}
	q(u_k^{(n)} = 1) \propto q(z^{(n)} | k) q(\bm{s}^{(n)} | k) q(k | \bm{x}^{(n)}) .
\end{align*}

\paragraph{[M-Step]}
In M-Step, we fix the distribution $q(u_k^{(n)} = 1) = \beta_k^{(n)}$, and maximize the lower-bound (\ref{eq:lb}) with respect to $w$, $\eta$, $\mu$, and $\lambda$.
The maximization over $w$ is a weighted multinomial logistic regression and can be solved efficiently, e.g. by using conjugate gradient method:
\begin{align*}
	\max_w \sum_{n=1}^N \sum_{k=1}^K \beta_k^{(n)} \log \frac{\exp (\bm{w}_k^\top \bm{s}^{(n)})}{\sum_{k'=1}^K \exp (\bm{w}_{k'}^\top \bm{s}^{(n)})} .
\end{align*}
The maximization over $\eta$ can be derived analytically as
\begin{align*}
	\eta_{k \ell} = \frac{\sum_{n=1}^N \beta_k^{(n) } s_\ell^{(n)}}{\sum_{n=1}^N \beta_k^{(n) }}
\end{align*}
The maximization over $\mu$ and $\lambda$ can also be derived as
\begin{align*}
	\mu_k = \frac{\sum_{n=1}^N \beta_k^{(n)} z^{(n)}}{\sum_{n=1}^N \beta_k^{(n)}} , \; \lambda_k = \frac{\sum_{n=1}^N \beta_k^{(n)}}{\sum_{n=1}^N \beta_k^{(n)} (z^{(n)} - \mu_k)^2} .
\end{align*}

\section{Experiments}
\label{sec:exp}

We evaluated the performance of the proposed method on a synthetic data and on an energy efficiency data~\cite{tsanas2012accurate}\footnote{http://archive.ics.uci.edu/ml/datasets/Energy+efficiency}.
For both datasets, we set $K = 4$.
As \emph{model P}, we used XGBoost~\cite{chen2016xgboost}.

We compared the proposed method with a decision tree.
We used \texttt{DecisionTreeRegressor} of scikit-learn in Python.
The optimal tree structure is selected from several candidates using 5-fold cross validation.

\subsection{Data Descriptions}
\label{sec:exp}

We prepared three i.i.d.\ datasets as $\mathcal{D}_{\rm ATM}$, $\mathcal{D}_{\rm train}$, and $\mathcal{D}_{\rm test}$, which are used for building an ATM, training \emph{model I}, and evaluating the quality of \emph{model I}, respectively.

\textbf{Synthetic Data:}
We generated regression data as follows:
\begin{align*}
	& \bm{x} = (x_1, x_2) \sim {\rm Uniform}[0, 1],  \\
	& y = {\rm XOR} (x_1 < 0.5, x_2 <0.5) + \epsilon, \;\; \epsilon \sim \mathcal{N}(0, 0.1^2) .
\end{align*}
For each of $\mathcal{D}_{\rm ATM}$, $\mathcal{D}_{\rm train}$, and $\mathcal{D}_{\rm test}$, we generated 1000 samples.

\textbf{Energy Efficiency Data:}
The energy efficiency dataset comprises 768 samples and eight numeric features, i.e., Relative Compactness, Surface Area, Wall Area, Roof Area, Overall Height, Orientation, Glazing Area, and Glazing Area Distribution.
The task is regression, which aims to predict the heating load of the building from these eight features.
In the experiment, we used $40\%$ of data points as $\mathcal{D}_{\rm ATM}$, another $30\%$ as $\mathcal{D}_{\rm train}$, and the remaining $30\%$ as $\mathcal{D}_{\rm test}$.

\subsection{Results}
\label{sec:res}

The found rules by the proposed method are shown in \figurename~\ref{fig:example}(c) (synthetic data), and in \tablename~\ref{tab:energy} (energy efficiency data).
On synthetic data, the proposed method could find the correct data structure as in \figurename~\ref{fig:example}(a).
On energy efficiency data, \tablename~\ref{tab:energy} shows that the found rules are easily interpretable and would be appropriate; the second and the third rules indicate that the predicted heating load $z$ is small when Relative Compactness is less than 0.75, while the last rule indicates that $z$ is large when Relative Compactness is more than 0.75.
These resulting rules are intuitive in that the load is small when the building is small, while the load is large when the building is huge.
Hence, from these simplified rules, we can infer that ATM is learned in accordance with our intuition about data.

\begin{figure}[t]
	\centering
	\includegraphics[width=0.31\textwidth]{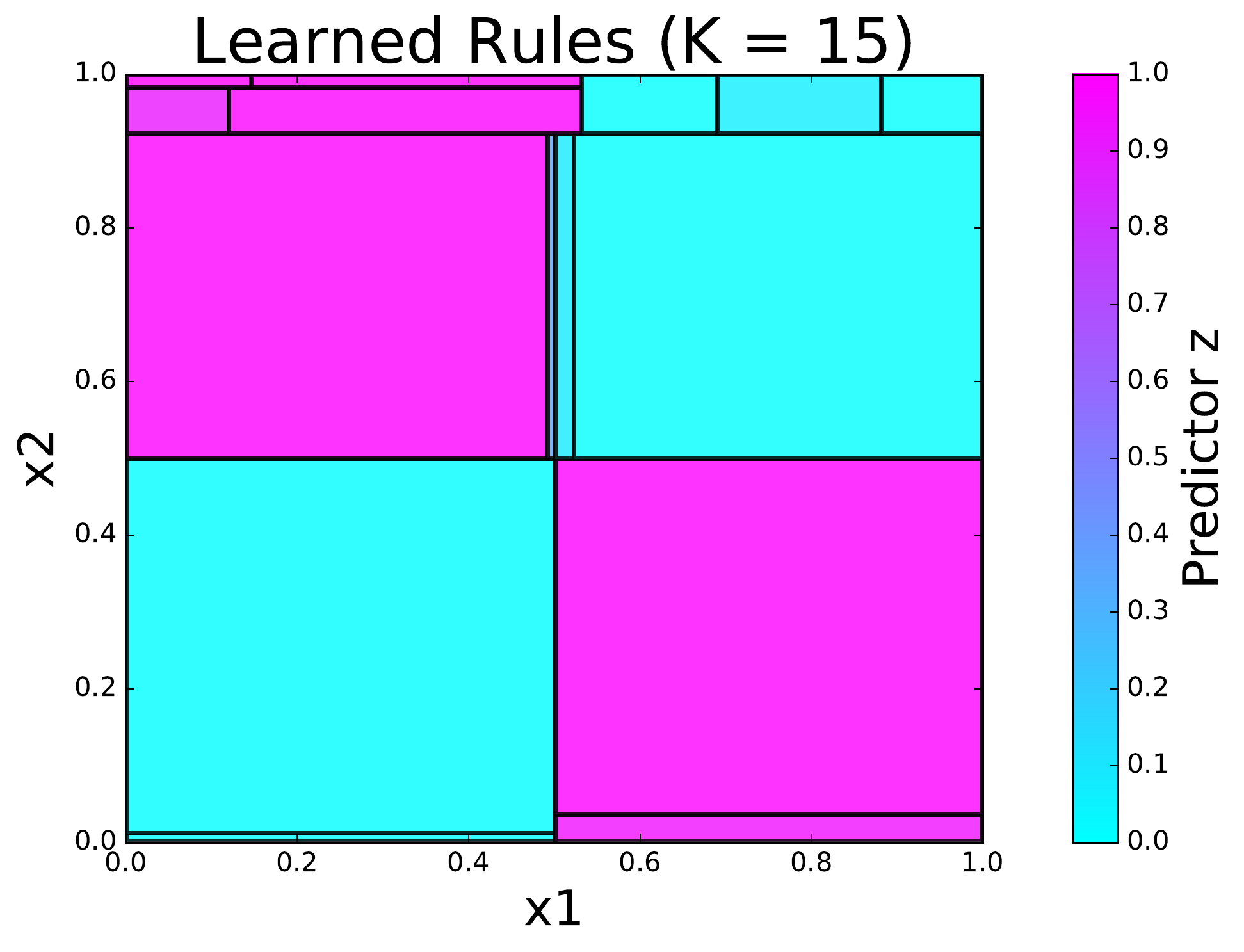}
	\caption{Synthetic Data: Decision Tree Rules}
	\label{fig:tree}
\end{figure}

\tablename~\ref{tab:error} summarizes the evaluation results of the proposed method and a decision tree.
On both data, the proposed method attained a reasonable prediction error using only $K=4$ rules.
In contrast, the decision tree tended to generate more than 10 rules.
On the synthetic data, the decision tree attained a smaller error than the proposed method while generating 15 rules which are nearly four times more than the proposed method (\figurename~\ref{fig:tree}).
On the energy efficiency data, the decision tree scored a significantly worse error indicating that the found rules may not be reliable, and thus inappropriate for the interpretation purpose.
From these results, we can conclude that the proposed method would be preferable for interpretation because it provided simple rules with reasonable predictive powers.

\begin{table}[t]
	\small
	\caption{Energy Efficiency Data: $K=4$ rules extracted using the proposed method, and three rules out of 37 rules found by the decision tree. $z$ is the predictor value corresponding to each rule.}
	\label{tab:energy}
	\centering
	\begin{tabular}{ccp{0.7\hsize}}
		& $z$ &  Rule\\
		\hline
		\multirow{6}{*}{\rotatebox[origin=c]{90}{Proposed}}
& $12.18$ & ${\rm Wall Area} < 281.75$, $0.20 \leq {\rm Glazing Area} < 0.33$, \\
& $13.41$ & ${\rm Relative Compactness} < 0.75$, \\
& $14.54$ & ${\rm Relative Compactness} < 0.75$, ${\rm Wall Area} < 330.75$, ${\rm Glazing Area} \geq 0.33$, \\
& $31.42$ & ${\rm Relative Compactness} \geq 0.75$, \\
		\hline
		\multirow{7}{*}{\rotatebox[origin=c]{90}{\shortstack{Decision\\Tree}}}
& $8.17$ & ${\rm Roof Area} < 5.25$, ${\rm Orientation} < 0.05$, \\
& $23.11$ & ${\rm Relative Compactness} < 624.75$, ${\rm Surface Area} < 306.25$, ${\rm Roof Area} \geq 5.25$, ${\rm Orientation} < 0.17$, ${\rm Glazing Area} < 2.50$, \\
& $41.32$ & ${\rm Relative Compactness} \geq 624.75$, ${\rm Roof Area} \geq 5.25$, ${\rm Overall Height} \geq 3.50$, ${\rm Orientation} \geq 0.32$, \\
	\end{tabular}
\end{table}

\begin{table}[t]
	\caption{Evaluation Results. The two measures are summarized as \emph{the number of rules} / \emph{test error}.}
	\label{tab:error}
	\centering
	\begin{tabular}{clll}
		& XGBoost & Proposed & Decision Tree\\
		\hline
		Synthetic & 744 / 0.01 & 4 / 0.02 &  15 / 0.01\\
		Energy & $>$100 / 0.22 & 4 / 20.19 &  37 / 168.19
	\end{tabular}
\end{table}

%
%

\section{Conclusion}
\label{sec:concl}

We proposed a post processing method that improves the interpretability of ATMs.
The difficulty of ATM interpretation comes from the fact that ATM divides an input space into more than a thousand of small regions.
We assumed that the model is interpretable if the number of regions is small.
Based on this principle, we formulated the ATM interpretation problem as an approximation of the ATM using a smaller number of regions.

There remains several open issues.
For instance, there is a freedom on the modeling (\ref{eq:q}).
While we adopted a fairly simple formulation, there may be another modeling that solves Problem~\ref{prob:atm} in a better way.
Another freedom is on the choice of a metric to measure the proximity between \emph{model P} and \emph{model I}.
We used KL divergence because we could derive an EM algorithm to learn \emph{model I} reasonably.
The choice of the number of regions $K$ also remains as an open issue.
Currently, we treat $K$ as a user tuning parameter, while, ultimately, it is desirable that the value of $K$ is determined automatically.


\bibliographystyle{icml2016}
\bibliography{postATM}

\end{document}